\documentclass[utf8]{FrontiersinHarvard} 

\usepackage{url,hyperref,lineno,microtype,subcaption}
\usepackage[onehalfspacing]{setspace}
\usepackage{multirow}
\usepackage{indentfirst}
\usepackage{bm}
\usepackage{color}
\usepackage{centeredline}

\def\keyFont{\fontsize{8}{11}\helveticabold }
\def\firstAuthorLast{Sample {et~al.}} 
\def\Authors{Yang Yang\,$^{1}$, Kaixiong Xu\, $^{1}$,Kaizheng Wang\,$^{2,*}$}
\def\Address{$^{1}$Faculty of Information Engineering and Automation, Kunming University of Science and Technology, Kunming, China\\
	$^{2}$Faculty of Electrical Engineering, Kunming University of Science and Technology, Kunming, China}
\def\corrAuthor{Kaizheng Wang}

\def\corrEmail{kz.wang@foxmail.com}

\begin{document}
	
	\onecolumn
	\firstpage{1}
	
	\title {Cascaded information enhancement and cross-modal attention feature fusion for multispectral pedestrian detection}
	
	\author[\firstAuthorLast ]{\Authors} 
	\address{} 
	\correspondance{} 
	
	\extraAuth{}

	\maketitle
	
	\noindent\begin{abstract}
		
		\section{}
		Multispectral pedestrian detection is a technology designed to detect and locate pedestrians in Color and Thermal images, which has been widely used in automatic driving, video surveillance, etc. So far most available multispectral pedestrian detection algorithms only achieved limited success in pedestrian detection because of the lacking take into account the confusion of pedestrian information and background noise in Color and Thermal images. Here we propose a multispectral pedestrian detection algorithm, which mainly consists of a cascaded information enhancement module and a cross-modal attention feature fusion module. On the one hand, the cascaded information enhancement module adopts the channel and spatial attention mechanism to perform attention weighting on the features fused by the cascaded feature fusion block. Moreover, it multiplies the single-modal features with the attention weight element by element to enhance the pedestrian features in the single-modal and thus suppress the interference from the background. On the other hand, the cross-modal attention feature fusion module mines the features of both Color and Thermal modalities to complement each other, then the global features are constructed by adding the cross-modal complemented features element by element, which are attentionally weighted to achieve the effective fusion of the two modal features. Finally, the fused features are input into the detection head to detect and locate pedestrians. Extensive experiments have been performed on two improved versions of annotations (sanitized annotations and paired annotations) of the public dataset KAIST. The experimental results show that our method demonstrates a lower pedestrian miss rate and more accurate pedestrian detection boxes compared to the comparison method. Additionally, the ablation experiment also proved the effectiveness of each module designed in this paper.
		\tiny
		\keyFont{ \section{Keywords:} multispectral pedestrian detection, attention mechanism, feature fusion, convolutional neural network, background noise} 
	\end{abstract}
	
	\section{Introduction}
	Pedestrian detection, parsing visual content to identify and locate pedestrians on an image/video, has been viewed as an essential and central task within the computer vision field and widely employed in various applications, e.g. autonomous driving, video surveillance and person re-identification \citep{jeong2017,zhang2018,lilingli2021,chenyiwen2021,dongneng2022,lishuang,wangshujuan}. The performance of such technology has greatly advanced through the facilitation of convolutional neural networks (CNN). Typically, pedestrian detectors take Color images as input and try to retrieve the pedestrian information from them. However, the quality of Color images highly depends on the light condition. Missing recognition of pedestrians occurs frequently when pedestrian detectors process Color images with poor resolution and contrast caused by unfavorable lighting. Consequently, the use of such models has been limited for the application of all-weather devices.
	
\begin{figure}[h!]
		\begin{center}
			\includegraphics[width=15cm]{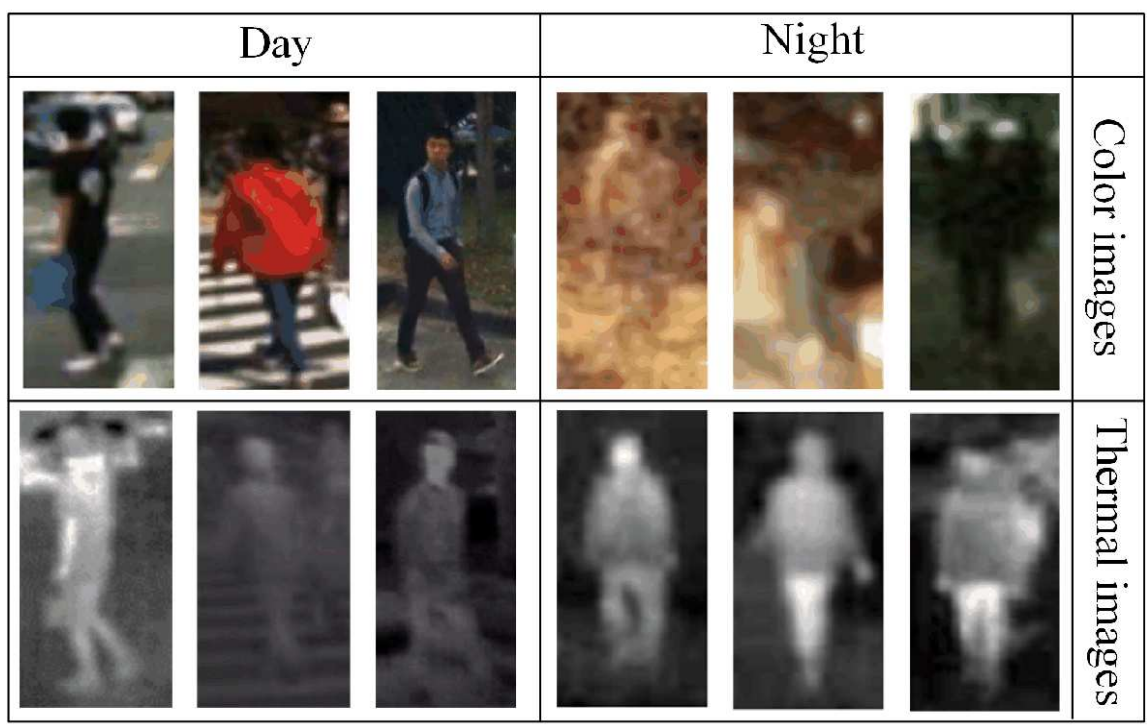}
		\end{center}
		\caption{Example of Color and Thermal images of pedestrians in daytime and nighttime scenes}\label{fig:1}
	\end{figure}

	Thermal imaging is related to the infrared radiation of pedestrians, barely affected by changes in ambient light. The technique of combining Color and Thermal images has been explored in recent years \citep{hwang2015,liu2016,gonzalez2016,yangmoyuan2020,liu2020,li2018joint,li2020discriminative,wangjiaxin2021,huangbochun}. These methods has been shown to exhibit positive effects on pedestrian detection performance in complex environments as it could retrieve more pedestrian information. However, despite important initial success, there remain two major challenges. First, as shown in Figure \ref{fig:1}, the image of pedestrians tends to blend with the background for nighttime Color images resulting from insufficient light \citep{zhuzhipin}, and for daytime Thermal images as well due to similar temperatures between the human body and the ambient environment \citep{yang2022}. Second, there is an essential difference between Color images and Thermal images the former displays the color and texture detail information of pedestrians while the latter shows the temperature information. Therefore, solutions needed to be taken to augment the pedestrian features in Color and Thermal modalities in order to suppress background interference, and enable better integration and understanding of both Color and Thermal images to improve the accuracy of pedestrian detection in complex environments.
	
	To address the challenges above, the researches \citep{guan2019,zhou2020} designed illumination-aware networks to obtain illumination-measured parameters of Color and Thermal images respectively, which were used as fusion weights for Color and Thermal features in order to realize a self-adaptively fuse of two modal features. However, the acquisition of illumination-measured parameters relied heavily on the classification scores, the accuracy of which was limited by the performance of the classifier. Li et al. \citep{li2022} reported confidence-aware networks to predict the confidence of detection boxes for each modal, and then Dempster-Sheffer theory combination rules were employed to fuse the results of different branches based on uncertainty. Nevertheless, the accuracy of predicting the detection boxes' confidence is also affected by the performance of the confidence-aware network. A cyclic fusion and refinement scheme was introduced by  \citep{zhang2020b} for the sake of gradually improving the quality of Color and Thermal features and automatically adjusting the complementary and consistent information balance of the two modalities to effectively utilize the information of both modalities. However, this method only used a simple feature cascade operation to fuse Color and Thermal features and failed to fully exploit the complementary features of these two modalities.
	
	To tackle the problems aforementioned, we propose a multispectral pedestrian detection algorithm with cascaded information enhancement and cross-modal attention feature fusion. The cascaded information enhancement module (CIEM) is designed to enhance the pedestrian information suppressed by the background in the Color and Thermal images. CIEM uses a cascaded feature fusion block to fuse Color and Thermal features to obtain fused features of both modalities. Since the fused features contain the consistency and complementary information of Color and Thermal modalities, the fused features can be used to enhance Color and Thermal features respectively to reduce the interference of background on pedestrian information. Inspired by the attention mechanism, the attention weights of the fused features are sequentially obtained by channel and spatial attention learning, and the Color and Thermal features are multiplied with the attention weights element by element, respectively. In this way, the single-modal features have the combined information of the two modalities, and the single-modal information is enhanced from the perspective of the fused features. Although CIEM enriches single-modal pedestrian features, simple feature fusion of the enhanced single-modal features is still insufficient for robust multispectral pedestrian detection. Thus, we design the cross-modal attention feature fusion module (CAFFM) to efficiently fuse Color and Thermal features. Cross-modal attention is used in this module to implement the differentiation of pedestrian features between different modalities. In order to supplement the pedestrian information of the other modality to the local modality, the attention of the other modality is adopted to augment the pedestrian characteristics of the local modality. A global feature is constructed by adding the Color and Thermal features after performing cross-modal feature enhancement, and the global feature is used to guide the fusion of the Color and Thermal features. Overall, the method presented in this paper enables more comprehensive pedestrian features acquisition through cascaded information enhancement and cross-modal attention feature fusion, which effectively enhances the accuracy of multispectral image pedestrian detection. The main contributions of this paper are summarized as follows:
	
	(1) A cascaded information enhancement module is proposed. From the perspective of fused features, it reduces the interference from the background of Color and Thermal modalities on pedestrian detection and augments the pedestrian features of Color and Thermal modalities separately through an attention mechanism.
	
	(2) The designed cross-modal attention feature fusion module first mines the features of both Color and Thermal modalities separately through a cross-modal attention network and adds them to the other modality for cross-modal feature enhancement. Meanwhile, the cross-modal enhanced Color and Thermal features are used to construct global features to guide the feature fusion of the two modalities.
	
	(3) Numerous experiments are conducted on the public dataset KAIST to demonstrate the effectiveness and superiority of the proposed method. In addition, the ablation experiments also demonstrate the effectiveness of the proposed modules.
	
	\section{RELATED WORKS}
	\subsection{Multispectral Pedestrian Detection}
	Multispectral sensors can obtain paired Color-Thermal images to provide complementary information about pedestrian targets. A large multispectral pedestrian detection (KAIST) dataset was constructed by  \citep{hwang2015}. Meanwhile, by combining the traditional aggregated channel feature (ACF) pedestrian detector   \citep{dollar2014} with the HOG algorithm  \citep{dalal2015}, an extended ACF (ACF+T+THOG) method was proposed to fuse Color and Thermal features. In 2016, Liu et al. \citep{liu2016} proposed four fusion modalities of low-layer feature, middle-layer feature, high-layer feature, and confidence fraction fusion with VGG16 as the backbone network, and the middle-layer feature fusion was proved to offer the maximum integration capability of Color and Thermal features. Inspired by this, \citep{konig2017} developed a multispectral region candidate network with Faster RCNN (Region with CNN features, RCNN) \citep{ren2017} as the architecture and replaced the original classifier in Faster RCNN with an enhanced decision tree classifier to reduce the missed and false detection of pedestrians. Recently, Kim et al. \citep{kim2021a} deployed the EfficientDet as the backbone network and proposed an EfficientDet-based fusion framework for multispectral pedestrian detection to improve the detection accuracy of pedestrians in Color and Thermal images by adding and cascading the Color and Thermal features. Although the studies \citep{hwang2015,liu2016,konig2017,kim2021a} fused Color and Thermal features for pedestrian detection, they mainly focused on exploring the impact of different stages of fusion on pedestrian detection, and only adopted simple feature fusion and not focusing on the case of pedestrian and background confusion.
	
	In 2019, Zhang et al. \citep{zhang2019a} observed a weak alignment problem of pedestrian position between Color and Thermal images, for which the KAIST dataset was re-annotated and Aligned Region CNN (AR-CNN) was proposed to handle weakly aligned multispectral pedestrian detection data in an end-to-end manner. But the deployment of this algorithm requires pairs of annotations, and the annotation of the dataset is a time-consuming and labor-intensive task, which makes the algorithm difficult to be applied in realistic scenes. Kim et al. \citep{kim2021b} proposed a new single-stage multispectral pedestrian detection framework. This framework used multi-label learning to learn input state-aware features based on the state of the input image pair by assigning an individual label (if the pedestrian is visible in only one image of the image pair, the label vector is assigned as $y_1 \in[0,1]$ or $ {y_2} \in [1,0] $ ; if the pedestrian is visible in both images of the image pair, the label vector is assigned as $ {y_3} \in [1,1] $ ) to solve the problem of weak alignment of pedestrian locations between Color and Thermal images, but the model still requires pairs of annotations during training. Guan et al. \citep{guan2019} designed illumination-aware networks to obtain illumination-measured parameters for Color and Thermal images separately and used them as the fusion weights for Color and Thermal features.  Zhou et al. \citep{zhou2020} designed a differential modality perception fusion module to guide the features of the two modalities to become similar, and then used the illumination perception network to assign fusion weights to the Color and Thermal features. Kim et al. \citep{kim2022} reported an uncertainty-aware cross-modal guidance (UCG) module to guide the distribution of modal features with high prediction uncertainty to align with the distribution of modal features with low prediction uncertainty. The researches \citep{guan2019,zhou2020} noticed that the pedestrians in Color and Thermal images are easily confused with the background and used illumination-aware networks to assign fusion weights to Color and Thermal features. However, the acquisition of illumination-measured parameters relied heavily on the classification scores, whose accuracy was limited by the performance of the classifier. In contrast, the method proposed in this paper not only considers the confusion of pedestrians and background in Color and Thermal images but also effectively fuses the two modal features.
	\subsection{Attention Mechanisms}
	Attention mechanisms \citep{atention} utilized in computer vision are aimed to perform the processing of visual information. Currently, attention mechanisms have been widely used in semantic segmentation \citep{li2020a}, image captioning  \citep{li2020b}, image fusion \citep{xiaowanxin2022,cenyueliang2021}, image dehazing \citep{gaojirui2022}, saliency target detection  \citep{xu2021}, person re-identification \citep{xukaixiong2022,ACMMM2021,wangyiming}, etc. Hu et al. \citep{hu2020} introduced the idea of a squeeze and excitation network (SENet) to simulate the interdependence between feature channels in order to generate channel attention to recalibrate the feature mapping of channel directions. Li et al. \citep{li2019a} employed the use of a selective kernel unit (SKNet) to adaptively fuse branches with different kernel sizes based on input information. A work inspired by this was from Dai et al.\citep{dai2021}. They designed a multi-scale channel attention feature fusion network that used channel attention mechanisms to replace simple fusion operations such as feature cascades or summations in feature fusion to produce richer feature representations. However, this recent progress in multispectral pedestrian detection has also been limited to two main challenges the interference caused by background and the difference of fundamental characteristics in Color and Thermal images. Therefore, we propose a multispectral pedestrian detection algorithm with cascaded information enhancement and cross-modal attention feature fusion based on the attention mechanism.
	\section{Methods}
\begin{figure}[h!]
		\begin{center}
			\includegraphics[width=18cm]{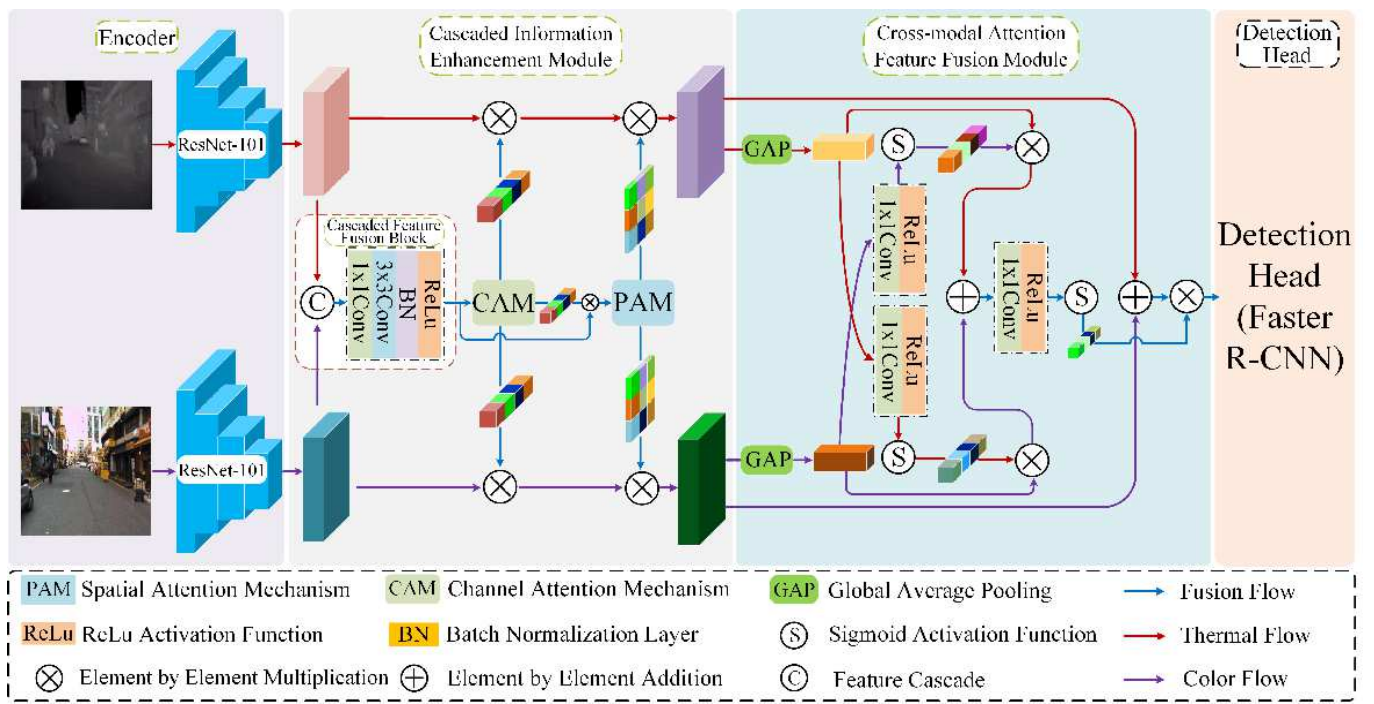}
		\end{center}
		\caption{Overall framework of the proposed algorithm}\label{fig:2}
	\end{figure}

	The overall network framework of the proposed algorithm is shown in Figure \ref{fig:2}. The network consists of an encoder, a cascaded information enhancement module (CIEM), a cross-modal attentional feature fusion module (CAFFM) and a detection head. Specifically, ResNet-101 \citep{he2016} is used as the backbone network of the encoder to encode the features of the input Color images $\boldsymbol{X}_c$ and Thermal images  $\boldsymbol{X}_t$  to obtain the corresponding feature maps ${\boldsymbol{F}_c} \in {{\rm{R}}^{W \times H \times C}}$  and  ${\boldsymbol{F}_t} \in {{\rm{R}}^{W \times H \times C}}$ ($ W $, $ H $,  $ C $ represent the width, height and the number of channels of the feature maps, respectively). CIEM enhances single-modal information from the perspective of fused features by cascading feature fusion blocks to fuse ${\boldsymbol{F}_c}$ and ${\boldsymbol{F}_t}$, and attention weighting the fused features to enrich pedestrian features. CAFFM complements the features of different modalities by mining the complementary features between the two modalities and constructs global features to guide the effective fusion of the two modal features. The detection head is employed for pedestrian recognition and localization of the final fused features.
	\subsection{Cascaded Information Enhancement Module}
	Considering the confusion of pedestrians with the backgrounds in Color and Thermal images, we design a cascaded information enhancement module (CIEM) to augment the pedestrian features of both modalities to mitigate the effect of background interference on pedestrian detection.
	Specifically, a cascaded feature fusion block is used to fuse the Color features  ${\boldsymbol{F}_c}$ and Thermal features ${\boldsymbol{F}_t}$ . The cascaded feature fusion block consists of feature cascade, $1 \times 1$ convolution, $3 \times 3$  convolution, $ BN$ layer, and $ReLu$ activation function. The feature cascade operation splice ${\boldsymbol{F}_c}$ and ${\boldsymbol{F}_t}$  along the direction of channels.  $1 \times 1$ convolution is conducive to cross-channel feature interaction in the channel dimension and reducing the number of channels in the splice feature map, while $3 \times 3$  convolution expands the field of perception and makes a more comprehensive fusion of features for generating fusion features $ {\boldsymbol{F}_{ct}}$ :
	\begin{equation}
		\boldsymbol{F}_{c t}={ReLu}\left({BN}\left({Conv}_3\left({Conv}_1\left[\boldsymbol{F}_c, \boldsymbol{F}_t\right]\right)\right)\right)
	\end{equation}
	where $ BN $ denotes batch normalization, $ Con{v_n}\left(  \cdot  \right)\ $ denotes a convolution kernel with kernel size $ n \times n $,
	$ [ \cdot , \cdot ] $ denotes the cascade of features along the channel direction,  $ ReLu( \cdot )$ represents $ReLu$   activation function. Fusion feature $ {\boldsymbol{F}_{ct}}$ is used to enhance the single-modal information because $ {\boldsymbol{F}_{ct}}$ combines the consistency and complementarity of the Color features  ${\boldsymbol{F}_c}$ and Thermal features ${\boldsymbol{F}_t}$ . The use of $ {\boldsymbol{F}_{ct}}$ for enhancing the single-modal feature can reduce the interference of the noise in the single-modal features (for example, it is difficult to distinguish between the pedestrian information and the background noise).
	
\begin{figure}[h!]
		\begin{center}
			\includegraphics[width=0.8\textwidth]{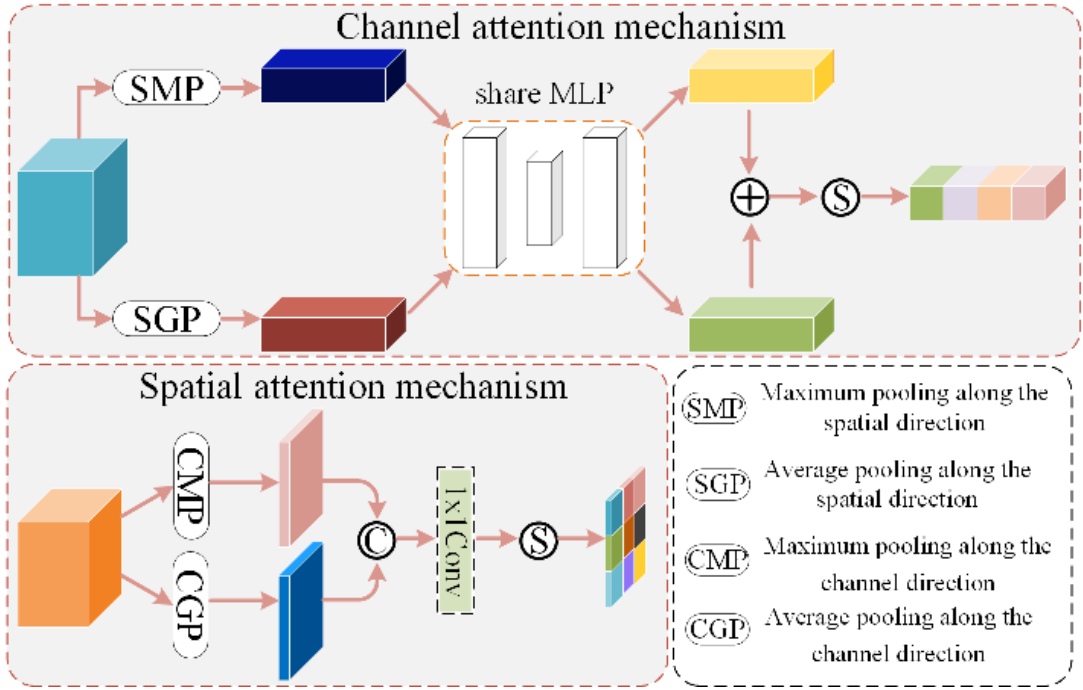}
		\end{center}
		\caption{Network structure of channel attention and spatial attention}\label{fig:3}
	\end{figure}

	In order to effectively enhance pedestrian features, the fusion feature $ {\boldsymbol{F}_{ct}}$ is sent into the channel attention module (CAM) and spatial attention module (PAM) \citep{woo2018} to make the network pay attention to pedestrian features. The network structure of CAM and PAM is shown in Figure \ref{fig:3}.  $ {\boldsymbol{F}_{ct}}$ first learns the channel attention weight ${\boldsymbol{w}_{ca}} \in {{\rm{R}}^{1 \times 1 \times C}}$  by CAM, then uses $ {\boldsymbol{w}_{ca}}$  to weight $ {\boldsymbol{F}_{ct}}$ , and the spatial attention weight $ {\boldsymbol{w}_{pa}} \in {{\rm{R}}^{W \times H \times 1}} $  is obtained from the weighted features by PAM.
	
	The single-modal Color features ${\boldsymbol{F}_c}$ and Thermal features  ${\boldsymbol{F}_t}$ are multiplied element by element with the attention weights  $ {\boldsymbol{w}_{ca}}$  and  $ {\boldsymbol{w}_{pa}}$  to enhance the single-modal features from the perspective of fused features. The whole process can be described as follows:
	\begin{equation}
		\boldsymbol{F}_t^{\prime}=\left(\boldsymbol{F}_t \otimes \boldsymbol{w}_{c a}\right) \otimes \boldsymbol{w}_{p a}
	\end{equation}
	\begin{equation}
		\boldsymbol{F}_c^{\prime}=\left(\boldsymbol{F}_c \otimes \boldsymbol{w}_{c a}\right) \otimes \boldsymbol{w}_{p a}
	\end{equation}
	where $\boldsymbol{F}_t^{\prime}$ and $\boldsymbol{F}_c^{\prime}$ denote the Color features and Thermal features obtained by the cascaded information enhancement module, respectively. $\otimes$ represents the element by element multiplication.
	\subsection{Cross-modal Attention Feature Fusion Module}
	There is an essential difference between Color and Thermal images, Color images reflect the color and texture detail information of pedestrians while Thermal images contain the temperature information of pedestrians, however, they also have some complementary information. In order to explore the complementary features of different image modalities and fuse them effectively, we design a cross-modal attention feature fusion module.
	
	Specifically, the Color features $\boldsymbol{F}_c^{\prime} $ and Thermal features $\boldsymbol{F}_t^{\prime}$ enhanced by CIEM are first mapped into feature vectors  $\boldsymbol{v}_c \in \mathrm{R}^{1 \times 1 \times C}$ and $\boldsymbol{v}_t \in \mathrm{R}^{1 \times 1 \times C}$ , respectively, by using global average pooling operation. The cross-modal attention network consists of a set of symmetric $1 \times 1$  convolutions,  $R e L u$ activation functions, and  $Sigmoid$ activation functions. In order to obtain the complementary features of the two modalities, more pedestrian features need to be mined from the single-modal. The feature vectors $\boldsymbol{v}_t$  and  $\boldsymbol{v}_c$ are learned to the respective modal attention weights $\boldsymbol{w}_t \in \mathrm{R}^{1 \times 1 \times C}$  and $\boldsymbol{w}_c \in \mathrm{R}^{1 \times 1 \times C}$  by a cross-modal attention network, and then the Color features  $\boldsymbol{F}_c^{\prime} $ are multiplied element by element with the attention weights $\boldsymbol{w}_t$  of the Thermal modality, and the Thermal features  $\boldsymbol{F}_t^{\prime}$ are multiplied element by element with the attention weights  $\boldsymbol{w}_c$  of the Color modality to complement the features of the other modality into the present modality. The specific process can be expressed as follows.
	\begin{equation}
		\boldsymbol{w}_t=\operatorname{\mathit{Sigmoid}}\left(\mathit{ReLu}\left(\mathit{Conv}_1\left(\mathit{GAP}\left(\boldsymbol{F}_t^{\prime}\right)\right)\right)\right)
	\end{equation}
	\begin{equation}
		\boldsymbol{F}_{c t}^{\prime}=\boldsymbol{w}_t \otimes G A P\left(\boldsymbol{F}_c^{\prime}\right)
	\end{equation}
	\begin{equation}
		\boldsymbol{w}_c={Sigmoid}\left({ReLu}\left({Conv}_1\left({GAP}\left(\boldsymbol{F}_c^{\prime}\right)\right)\right)\right)
	\end{equation}
	\begin{equation}
		\boldsymbol{F}_{t c}^{\prime}=\boldsymbol{w}_c \otimes G A P\left(\boldsymbol{F}_t^{\prime}\right)
	\end{equation}
	where  $\boldsymbol{F}_{c t}^{\prime}$ denotes Color features after supplementation with Thermal features, $\boldsymbol{F}_{t c}^{\prime}$  denotes Thermal features after supplementation with Color features, $G A P(\cdot)$  denotes global average pooling operation, ${Conv}_1(\cdot)$  denotes convolution with convolution kernel size $1 \times 1$ , $ ReLu( \cdot )$ denotes   $ReLu$ activation operation, and $Sigmoid$ $(\cdot)$  denotes $Sigmoid $  activation operation.
	
	In order to efficiently fuse the two modal features, the features $\boldsymbol{F}_{c t}^{\prime}$  and $\boldsymbol{F}_{t c}^{\prime}$  are subjected to an element by element addition operation to obtain a global feature vector containing Color and Thermal features. Then, the features $\boldsymbol{F}_t^{\prime}$ and  $\boldsymbol{F}_c^{\prime} $ are added element by element and multiplied with the attention weight $\boldsymbol{w}_{c t}$  of the global feature vector element by element to guide the fusion of Color and Thermal features from the perspective of global features to obtain the final fused feature $\boldsymbol{F}$. The fused feature $\boldsymbol{F}$ is input to the detection head to obtain the pedestrian detection results. The feature fusion process can be expressed as follows:
	\begin{equation}
		\boldsymbol{w}_{c t}={Sigmoid}\left({ReLu}\left({Conv}_1\left(\boldsymbol{F}_{c t}^{\prime} \oplus \boldsymbol{F}_{t c}^{\prime}\right)\right)\right)
	\end{equation}
	\begin{equation}
		\boldsymbol{F}=\boldsymbol{w}_{c t} \otimes\left(\boldsymbol{F}_t^{\prime} \oplus \boldsymbol{F}_c^{\prime}\right)
	\end{equation}
	where  $\oplus$ denotes element by element addition.
	\subsection{Loss Function}
	The loss function in this paper is consistent with the literature \citep{ren2017} and uses the Region Proposal Network (RPN) loss function $L_{R P N}$  and Fast RCNN \citep{girshick2015} loss function $L_{F R}$  to jointly optimize the network:
	\begin{equation}
		L=L_{R P N}+L_{F R}
	\end{equation}
	
	Both $L_{R P N}$ and $L_{F R}$ consist of classification loss $L_{c l s}$ and bounding box regression loss $L_{r e g}$:
	\begin{equation}
		L\left(\left\{p_i\right\},\left\{t_i\right\}\right)=\frac{1}{N_{c l s}} \sum_i L_{c l s}\left(p_i, p_i^*\right)+\lambda \frac{1}{N_{\text {reg }}} \sum_i p_i^* L_{r e g}\left(t_i, t_i^*\right)
	\end{equation}
	Where, $N_{c l s}$ is the number of anchors, $N_{r e g}$ is the sum of positive and negative sample number, $p_{i}$ is the probability that the $i$-th anchor is predicted to be the target, $p_i^*$ is 1 when the anchor is a positive sample, otherwise it is 0, $t_{i}$ denotes the bounding box regression parameter predicting the $i$-th anchor, and $t_i^*$ denotes the GT bounding box parameter of the $i$-th anchor, $\lambda=1$.
	
	The difference between the classification loss of RPN network and Fast RCNN network is that the RPN network focuses only on the foreground and background when classifying, so its loss is a binary cross-entropy loss, while the Fast RCNN classification is focused to the target category and is a multi-category cross-entropy loss:
	\begin{equation}
		L_{c l s}\left(p_i, p_i^*\right)=-\log \left[p_i^* p_i+\left(1-p_i^*\right)\left(1-p_i\right)\right]
	\end{equation}
	
	The bounding box regression loss of RPN network and Fast RCNN network uses Smooth $_{L_1}$ loss:
	\begin{equation}
		L_{\text {reg }}\left(t_i, t_i^*\right)=R\left(t_i-t_i^*\right)
	\end{equation}
	Where, R denotes Smooth $_{L_1}$ function,
	\begin{equation}
		\text { Smooth }_{L_1}(x)=\left\{\begin{array}{cc}
			\frac{x^2}{2 \sigma^2} & \text { if }|x|<\frac{1}{\sigma^2} \\
			|x|-0.5 & \text { otherwise }
		\end{array}\right.
	\end{equation}
	
	The difference between the bounding box regression loss of RPN loss and the regression loss of Fast RCNN loss is that the RPN network is trained when $\sigma$ =3 and the Fast RCNN network is trained when $\sigma$ =1.
	\section{EXPERIMENTAL RESULTS AND ANALYSIS}
	\subsection{Datasets}
	This paper evaluates the algorithm performance on the KAIST pedestrian dataset \citep{hwang2015}, which is composed of 95,328 pairs of Color and Thermal images captured during daytime and nighttime. It is the most widely used multispectral pedestrian detection dataset at present. The dataset is labeled with four categories including person, people, person?, and cyclist. Considering the application areas of multispectral pedestrian detection (e.g., automatic driving), all four categories are treated as positive examples for detection in this paper. To address the problem of the annotation errors and missing annotations in the original annotation of the KAIST dataset, studies \citep{liu2016,li2018,zhang2019a} performed data cleaning and re-annotation of the original data. Given that the annotations used in various studies are not consistent, we use 7601 pairs of Color and Thermal images from synthetic annotation (SA) \citep{li2018} and 8892 pairs of Color and Thermal images from paired annotation (PA) \citep{zhang2019a} for model training. The test set consists of 2252 pairs of Color and Thermal images, of which 1455 pairs are from the daytime and 797 pairs are from the nighttime. For a fair comparison with other methods, the test experiments were performed according to the reasonable settings proposed in the literature \citep{hwang2015}.
	\subsection{Evaluation Indexes}
	In this paper, Log-average Miss Rate (MR) proposed by Dollar et al.\citep {dollar2012} is employed as an evaluation index and combined with the plotting of the Miss Rate-FPPI curve to assess the effectiveness of the algorithm. The horizontal coordinate of the Miss Rate-FPPI curve indicates the average number of False Positives Per Image (FPPI), and the vertical coordinate represents the Miss Rate (MR), which is expressed as:
	\begin{equation}
		\text { MissRate }=\frac{F N}{T P+F N}
	\end{equation}
	\begin{equation}
		F P P I=\frac{F P}{\text { Total }(\text { images })}
	\end{equation}
	where ${F N}$ denotes False Negative, ${T P}$ denotes True Positive, ${F P}$ denotes False Positive, the sum of ${T P}$ and ${F N}$ is the number of all positive samples, and $\text { Total }(\text { images })$  denotes the total number of predicted images. It is worth noting that the lower the Miss Rate-FPPI curve trend, the better the detection performance; the smaller the MR value, the better the detection performance.
	In order to calculate MR, in logarithmic space, 9 points are taken from the horizontal coordinate (limited value range is $\left[10^{-2}, 10^0\right]$ ) of Miss Rate-FPPI curve, and then there are 9 corresponding vertical coordinates $m_1$, $m_2$,...$m_9$ . By averaging these values, MR can be obtained as follows:
	\begin{equation}
		\mathrm{MR}=\exp \left[\frac{1}{n} \sum_{i=1}^n \ln \left(m_i\right)\right]
	\end{equation}
	where $n$ is 9.
	\subsection{Implementation Details}
	In this paper, the deep learning framework pytorch1.7 is adopted. The experimental platform is the ubuntu18.04 operating system and a single NVIDIA GeForce RTX 2080Ti GPU. Stochastic Gradient Descent (SGD) algorithm is used to optimize the network during model training, with momentum value of 0.9, weight attenuation value $5 \times 10^{-4}$ , and initial learning rate is $1 \times 10^{-3}$ . The model is iterated for 5 epochs with the batch size of 4, and the learning rate decay to $1 \times 10^{-4}$  after the 3rd epoch.
	\subsection{Experimental Results and Analysis}
	\subsubsection{Construction of the Baseline}
	This work constructs a baseline algorithm architecture based on ResNet-101 backbone network and Faster RCNN detection head. Simple characteristic fusion (feature cascade, element by element addition and element by element multiplication) of the Color and Thermal features output by the backbone network is carried out in three sets of experiments. The fused feature is used as the input of the detection head. In order to ensure the high efficiency of the build baseline algorithm, synthesis annotation is employed to train and test the baseline. The test results are shown in Table \ref{tab:1}. The MR values using feature cascade, element by element addition and element by element multiplication in the all-weather scene are 14.62$\%$, 13.84$\%$ and 14.26$\%$, respectively. By comparing these three results, it can be seen that the feature element by element addition demonstrates the best performance. Therefore, we adopt the method of adding features element by element as the baseline integration method.
	\begin{table}[!ht]
		\centering
		\begin{spacing}{1.5}
			\caption{Experimental results of baseline under different fusion modes}\label{tab:1}
			\begin{tabular}{cc} \hline
				Fusion modes & All-weather  \\ \hline
				feature cascade & 14.62   \\
				element by element multiplication &	14.26  \\
				element by element addition & \textbf{\scriptsize 13.84}   \\ \hline
			\end{tabular}
		\end{spacing}
	\end{table}
	\subsubsection{Performance comparison of different methods}
	The performance of this method is compared with several other state-of-the-art methods. The compared methods include hand-represented methods, e.g., ACT+T+THOG \citep{hwang2015} and deep learning-based methods, e.g., Halfway Fusion \citep{liu2016}, CMT\_CNN\citep{xu2017}, CIAN\citep{zhang2019b}, IAF R-CNN\citep{li2019b}, IATDNN+IAMSS\citep{guan2019}, CS-RCNN \citep{zhang2020a}, IT-MN \citep{zhuang2022}, and DCRD \citep{liu2022}. Here, the model is trained using 7601 pairs of Color and Thermal images from SA and 8892 pairs of Color and Thermal images from PA, respectively. Besides, 2252 pairs of Color and Thermal images from the test set are used for model testing. Table \ref{tab:2} lists the experimental results.
	
	Table \ref{tab:2} shows that when the model is trained with SA, the MRs of the method proposed in this paper are 10.71$\%$, 13.09$\%$ and 8.45$\%$ for all-weather, daytime and nighttime scenes, respectively, which are 0.72$\%$, -1.23$\%$ and 0.37$\%$ lower than the compared method CS-RCNN with the best performance, respectively. The PA (Color) and PA (Thermal) in Table \ref{tab:2} represent the Color annotation and Thermal annotation in the pairwise annotation PA, respectively, for the purpose of training the model. It can be seen from \ref{tab:2} that the MRs of the method in this paper are 11.11$\%$ and 10.98$\%$ when using Color annotation and Thermal annotation in the all-weather scene, which are 2.53$\%$ and 3.70$\%$, respectively, lower than those of compared method with the best performance. In addition, by analyzing the experimental results of two improved versions of annotations, it can be found that pedestrian detection results are different when using different annotations, indicating the importance of annotations.
	\begin{table}[!ht]
		\centering
		\setlength{\tabcolsep}{1mm}
		\begin{spacing}{1.5}
			\caption{MRs of different methods on KAIST datasets}\label{tab:2}
			\begin{tabular}{cccccccccc}\hline
				\multicolumn{1}{c}{\multirow{2}{*}{Methods}} & \multicolumn{3}{c}{SA} & \multicolumn{3}{c}{PA(Color)} & \multicolumn{3}{c}{PA(Thermal)} \\ \cline{2-10}
				& All-weather  & Day & Night & All-weather  & Day & Night & All-weather  & Day & Night  \\ \hline
				ACF+T+THOG& 41.65 & 39.18 & 48.29 & 41.74 & 39.30 & 49.52 & 41.36 & 38.74 & 48.30\\
				Halfway Fusion& 25.75 & 24.88 & 26.59 & 25.10 & 24.29 & 26.12 & 25.51 & 25.20 & 24.90\\
				CMT\_CNN& 36.83 & 34.56 & 41.82 & 36.25 & 34.12 & 41.21 & -- & -- & --\\
				IAF R-CNN& 15.73 & 14.55 & 18.26 & 15.65 & 14.95 & 18.11 & 16.00 & 15.22 & 17.56\\
				IATDNN+IAMSS& 14.95 & 14.67 & 15.72 & 15.14 & 14.82 & 15.87 & 15.08 & \textcolor{blue}{15.02} & 15.20\\
				CIAN& 14.12 & 14.77 & 11.13 & 14.64& 15.13 & \textcolor{blue}{12.43} & \textcolor{blue}{14.68} & 16.21 & \textcolor{blue}{9.88}\\
				CS-RCNN & \textcolor{blue}{11.43} & \textcolor{red}{11.86} & \textcolor{blue}{8.82} & -- & -- & -- & -- & -- & -- \\
				IT-MN& 14.19 & 14.30 & 13.98 & -- & -- & -- & -- & -- & -- \\
				DCRD& 12.58 & 13.12 & 11.65 & \textcolor{blue}{13.64} & \textcolor{blue}{13.15} & 13.98 & -- & -- & --\\
				Ours& \textcolor{red}{10.71} & \textcolor{blue}{13.09} & \textcolor{red}{8.45} & \textcolor{red}{11.11} & \textcolor{red}{12.85} & \textcolor{red}{8.77} & \textcolor{red}{10.98} & \textcolor{red}{13.07} & \textcolor{red}{8.53} \\ \hline
			\end{tabular}
		\end{spacing}
	\end{table}
	\subsubsection{Analysis of Ablation Experiments}
	\noindent(1) Complementarity and importance of Color and Thermal features
	\begin{table}[!ht]
		\centering
		\begin{spacing}{1.5}
			\caption{MRs of different modal inputs}\label{tab:3}
			\begin{tabular}{cccc} \hline
				Input & All-weather& Day & Night \\ \hline
				dual-stream Color images & 25.37& 19.31  & 31.18\\
				dual-stream Thermal images &	17.55 & 22.81 & 12.61 \\
				Color images + Thermal images & \textbf{\scriptsize 13.84} & \textbf{\scriptsize 15.35} & \textbf{\scriptsize 12.48} \\ \hline
			\end{tabular}
		\end{spacing}
	\end{table}
	
	This section compares the effect of different input sources on pedestrian detection performance. In order to eliminate the impact of the proposed module on detection performance, three sets of experiments are conducted on baseline: 1) the combination of Color and Thermal images as the input source (the input of the two branches of the backbone network are respectively Color and Thermal images); 2) dual-stream Color image as the input source (use Color images to replace Thermal images, that is, the backbone network input source is Color images); 3) dual-stream Thermal images as the input source (use Thermal images to replace Color images, that is, the backbone network input source is Thermal images).The training set of the model here is 7061 pairs of images of SA, and the test set is 2252 pairs of Color and Thermal images. Table \ref{tab:3} shows the MRs of these three input sources for the all-weather, daytime, and nighttime scenes. It can be seen from Table \ref{tab:3} that the MRs obtained using Color and Thermal images as input to the network are 13.84$\%$, 15.35$\%$ and 12.48$\%$ for the all-weather, daytime and nighttime scenes, respectively, which are 11.53$\%$, 3.96$\%$, 18.70$\%$ and 3.71$\%$, 7.46$\%$, 0.13$\%$ lower than using Color images and Thermal images as input alone. The experimental results prove that the detection network combining Color and Thermal features delivers better performance, indicating that Color and Thermal features are important for pedestrian detection.
	
\begin{figure}[h!]
		\begin{center}
			\includegraphics[width=18cm]{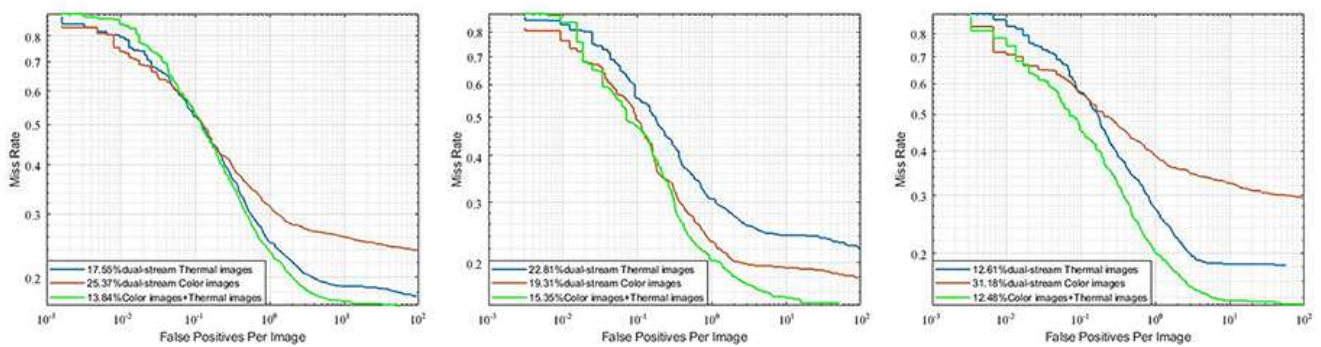}
		\end{center}
		\caption{The Miss Rate-FPPI curves of the detection results of the three groups of input sources in the All-weather, Daytime and Nighttime scenes (From left to right, All-weather, Daytime and Nighttime Miss Rate-FPPI curves are shown in the figure)}\label{fig:4}
	\end{figure}

	Figure \ref{fig:4} shows the Miss Rate-FPPI curves of the detection results for these three input sources in the all-weather, daytime, and nighttime scenes (blue, red and green curves indicate dual-stream Thermal images, dual-stream Color images, and Color and Thermal images, respectively). By analyzing the Miss Rate-FPPI curve trend and combining with the experimental data in Table \ref{tab:3}, it can be seen that the detection effect of Color images as the input source is better than that of Thermal images in the daytime scene while the result is the opposite for the night scene, and the detection effect of Color and Thermal images combined as the input source is better than that of single-modal input in both daytime and nighttime. It shows that there are complementary features between Color and Thermal modalities, and the fusion of the two modal features can improve the pedestrian detection performance.\\
	
	\noindent(2) Ablation experiments
	
	In this section, ablation experiments are conducted to demonstrate the effectiveness of the proposed cascaded information enhancement module (CIEM) and cross-modal attentional feature fusion module (CAFFM). Here, 7061 pairs of SA images are used to train the model, and 2252 pairs of Color and Thermal images in the test set are used to test the model.
	
	Effectiveness of CIEM: CIEM is used to enhance the pedestrian features in Color and Thermal images to reduce the interference from the background. The experimental results are shown in Table \ref{tab:4}. The MRs of baseline on SA are 13.84$\%$, 15.35$\%$ and 12.48$\%$ for all-weather, daytime and nighttime scenes, respectively. When CIEM is additionally employed, the MRs are 11.21$\%$, 13.15$\%$ and 9.07$\%$ for all-weather, daytime and nighttime scenes, respectively, which are reduced by 2.63$\%$, 2.20$\%$ and 3.41$\%$ compared to the baseline, respectively. It is shown that the proposed CIEM effectively enhances the pedestrian features in both modalities, reduces the interference of background, and improves the pedestrian detection performance.
	\begin{table}[!ht]
		\centering
		\begin{spacing}{1.5}
			\caption{MRs for ablation studies of the proposed method on SA}\label{tab:4}
			\begin{tabular}{cccc} \hline
				Methods & All-weather& Day & Night \\ \hline
				baseline & 13.84 & 15.35  & 12.48\\
				baseline + CIEM &	11.21 & 13.15 & 9.07 \\
				baseline + CAFFM & 11.68 & 13.81 & 9.50 \\
				Overall model & \textbf{\scriptsize 10.71} & \textbf{\scriptsize 13.09} & \textbf{\scriptsize 8.45} \\ \hline
			\end{tabular}
		\end{spacing}
	\end{table}
	
	Validity of CAFFM: CAFFM is used to effectively fuse Color and Thermal features. The experimental results are shown in Table \ref{tab:4}. On the SA, when the baseline is used with CAFFM, the MRs are 11.68$\%$, 13.81$\%$ and 9.50$\%$ in all-weather, daytime and nighttime scenes, respectively, which are reduced by 2.16$\%$, 1.54$\%$ and 2.98$\%$ compared baseline, respectively. It shows that the proposed CAFFM effectively fuses the two modal features to achieve robust multispectral pedestrian detection.
	
	Overall effectiveness: The proposed CIEM and CAFFM are additionally used on the basis of baseline. Experimental results show a reduction of 3.13$\%$, 2.26$\%$ and 4.03$\%$ in MRs for all-weather, daytime and nighttime scenes, respectively, compared to the baseline, indicating the overall effectiveness of the proposed method. A closer look reveals that with additional employment of CIEM and CAFFM alone, MRs are decreased by 2.63$\%$ and 2.16$\%$, respectively, in the all-weather scene, but the MR of the overall model is reduced by 3.13$\%$. It demonstrates that there is some orthogonal complementarity in the role of the proposed two modules.
	
\begin{figure}[h!]
		\begin{center}
			\includegraphics[width=18cm]{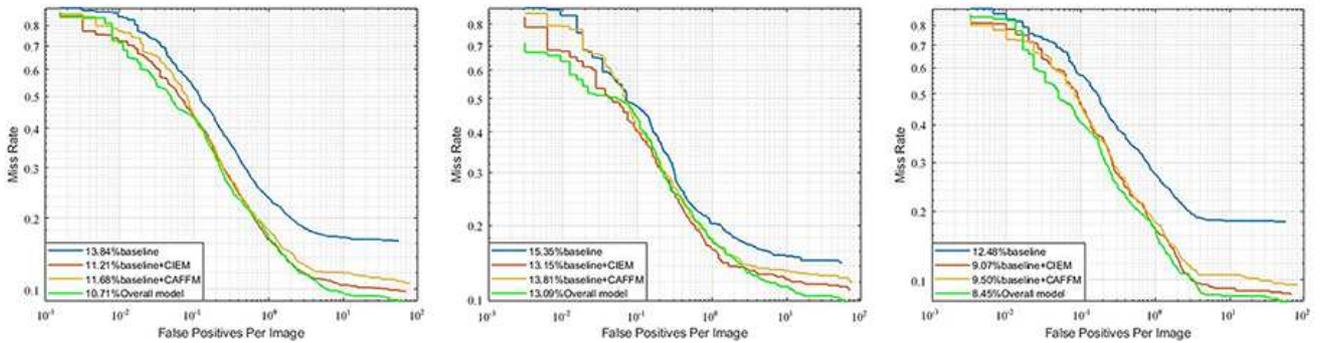}
		\end{center}
		\caption{The Miss Rate-FPPI curves of CIEM and CAFFM ablation studies in All-weather, Daytime and Nighttime scenes (From left to right, All-weather, Daytime and Nighttime Miss Rate-FPPI curves are shown in the figure)}\label{fig:5}
	\end{figure}

	Figure \ref{fig:5} shows the Miss Rate-FPPI curves for CIEM and CAFFM ablation studies in all-weather, daytime and nighttime scenes (blue, red, orange and green curves represent baseline, baseline + CIEM, baseline + CAFFM and overall model, respectively). It is clear that the curve trends of each module and the overall model are both lower than that of the baseline, which further proves the effectiveness of the method presented in this work.
	
\begin{figure}[h!]
		\begin{center}
			\includegraphics[width=18cm]{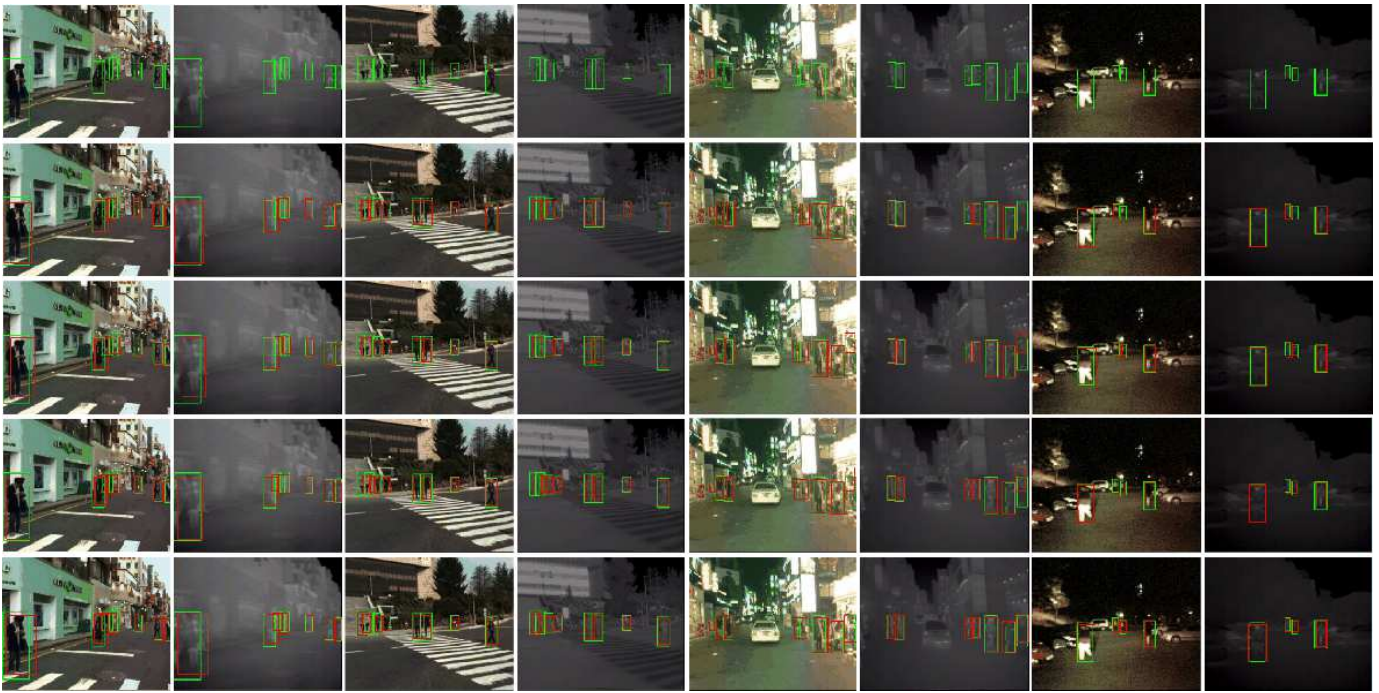}
		\end{center}
		\caption{In this paper, each module and baseline pedestrian detection results (The first row is the visualization results of labeled boxes for Color and Thermal images, and the second to the fifth rows are the visualization results of the labeled and prediction boxes for baseline, baseline + CIEM, baseline + CAFFM and the overall model pedestrian detection with the green and red boxes representing the labeled and prediction boxes, respectively.)}\label{fig:6}
	\end{figure}

	Furthermore, in order to qualitatively analyze the effectiveness of the proposed CIEM and CAFFM, four pairs of Color and Thermal images (two pairs of images are taken from daytime and two pairs of images are taken from nighttime) are selected from the test set for testing. The pedestrian detection results of the baseline and each proposed module are shown in Figure \ref{fig:6}. The first row is the visualization results of labeled boxes for Color and Thermal images, and the second to the fifth rows are the visualization results of the labeled and prediction boxes for baseline, baseline + CIEM, baseline + CAFFM, and the overall model pedestrian detection with the green and red boxes representing the labeled and prediction boxes, respectively. It can be seen that the proposed method successfully addresses the problem of pedestrian missing detection in complex environments and achieves more accurate detection boxes. For example, the second row, pedestrian detection missing happens in the first, third, and fourth pairs of images in the baseline detection result, however, the pedestrian miss detection problem is properly solved with CIEM and CAFFM added to the baseline and the overall model produces more accurate pedestrian detection boxes.
	\section{CONCLUSION}
	In this paper, we propose a multispectral pedestrian detection algorithm including cascaded information enhancement module and cross-modal attention feature fusion module. The proposed method improves the accuracy of pedestrian detection in multispectral images (Color and Thermal images) by effectively fusing the features from the two modules and augmenting the pedestrian features. Specifically, on the one hand, a cascaded information enhancement module (CIEM) is designed to enhance single-modal features to enrich the pedestrian features and suppress interference from the background noise. On the other hand, unlike previous methods that simply splice Color and Thermal features directly, a cross-modal attention feature fusion module (CAFFM) is introduced to mine the features of both Color and Thermal modalities and to complement each other, then complementary enhanced modal features are used to construct global features. Extensive experiments have been conducted on two improved annotations of the public dataset KAIST. The experimental results show that the proposed method is conducive to obtain more comprehensive pedestrian features and improve the accuracy of multispectral image pedestrian detection.
	
	\section{Tables}

	\section*{Conflict of Interest Statement}
	The authors declare that the research was conducted in the absence of any commercial or financial relationships that could be construed as a potential conflict of interest.
	
	\section*{Author Contributions}
	YY responsible for paper scheme design, experiment and paper writing. WK guide the paper scheme design and revision. XK guide to do experiments and write papers.
	
	\section*{Funding}
	This work was supported by the National Natural Science Foundation of China (No. 52107017) and Fundamental Research Fund of Science and Technology Department of Yunnan Province(No.202201AU070172)
	
	
	\bibliographystyle{Frontiers-Vancouver} 
	\bibliography{test}

\begin{thebibliography}{54}
\expandafter\ifx\csname natexlab\endcsname\relax\def\natexlab#1{#1}\fi
\expandafter\ifx\csname urlstyle\endcsname\relax
  \expandafter\ifx\csname doi\endcsname\relax
  \def\doi#1{doi:\discretionary{}{}{}#1}\fi \else
  \expandafter\ifx\csname doi\endcsname\relax
  \def\doi{doi:\discretionary{}{}{}\begingroup \urlstyle{rm}\Url}\fi \fi
\expandafter\ifx\csname selectlanguage\endcsname\relax
  \def\selectlanguage#1{}\fi

\bibitem[{Jeong et~al.(2017)Jeong, Ko, and Nam}]{jeong2017}
Jeong M, Ko BC, Nam JY.
\newblock Early detection of sudden pedestrian crossing for safe driving during
  summer nights.
\newblock {\em IEEE Transactions on Circuits and Systems for Video
  Technology\/} {\bf 27} (2017) 1368--1380.
\newblock \doi{10.1109/TCSVT.2016.2539684}.

\bibitem[{Zhang et~al.(2018)Zhang, Cheng, Gong, Shi, Qiu, Xia
  et~al.}]{zhang2018}
Zhang S, Cheng D, Gong Y, Shi D, Qiu X, Xia Y, et~al.
\newblock Pedestrian search in surveillance videos by learning discriminative
  deep features.
\newblock {\em Neurocomputing\/} {\bf 283} (2018) 120--128.

\bibitem[{Li et~al.(2021a)Li, Xie, Li, Zhang, Li, and Tan}]{lilingli2021}
Li L, Xie M, Li F, Zhang Y, Li H, Tan T.
\newblock Unsupervised domain adaptive person re-identification guided by
  low-rank priori.
\newblock {\em Journal of Chongqing University\/} {\bf 44} (2021a) 57--70.
\newblock \doi{10.11835/j.issn.1000-582X.2021.11.008}.

\bibitem[{Li et~al.(2021b)Li, Chen, Tao, Yu, and Qi}]{chenyiwen2021}
Li H, Chen Y, Tao D, Yu Z, Qi G.
\newblock Attribute-aligned domain-invariant feature learning for unsupervised
  domain adaptation person re-identification.
\newblock {\em IEEE Transactions on Information Forensics and Security\/} {\bf
  16} (2021b) 1480--1494.
\newblock \doi{10.1109/TIFS.2020.3036800}.

\bibitem[{Li et~al.(2022{\natexlab{a}})Li, Dong, Yu, Tao, and
  Qi}]{dongneng2022}
Li H, Dong N, Yu Z, Tao D, Qi G.
\newblock Triple adversarial learning and multi-view imaginative reasoning for
  unsupervised domain adaptation person re-identification.
\newblock {\em IEEE Transactions on Circuits and Systems for Video
  Technology\/} {\bf 32} (2022{\natexlab{a}}) 2814--2830.
\newblock \doi{10.1109/TCSVT.2021.3099943}.

\bibitem[{Li et~al.(2022{\natexlab{b}})Li, Li, Wang, Qi, and Li}]{lishuang}
Li S, Li F, Wang K, Qi G, Li H.
\newblock Mutual prediction learning and mixed viewpoints for
  unsupervised-domain adaptation person re-identification on blockchain.
\newblock {\em Simulation Modelling Practice and Theory\/} {\bf 119}
  (2022{\natexlab{b}}) 102568.

\bibitem[{Wang et~al.(2023)Wang, Liu, Li, Qi, and Yu}]{wangshujuan}
Wang S, Liu R, Li H, Qi G, Yu Z.
\newblock Occluded person re-identification via defending against attacks from
  obstacles.
\newblock {\em IEEE Transactions on Information Forensics and Security\/} {\bf
  18} (2023) 147--161.
\newblock \doi{10.1109/TIFS.2022.3218449}.

\bibitem[{Hwang et~al.(2015)Hwang, Park, Kim, Choi, and Kweon}]{hwang2015}
Hwang S, Park J, Kim N, Choi Y, Kweon IS.
\newblock Multispectral pedestrian detection: Benchmark dataset and baseline.
\newblock {\em 2015 IEEE Conference on Computer Vision and Pattern Recognition
  (CVPR)\/} (2015), 1037--1045.
\newblock \doi{10.1109/CVPR.2015.7298706}.

\bibitem[{Liu et~al.(2016)Liu, Zhang, Wang, and Metaxas}]{liu2016}
Liu J, Zhang S, Wang S, Metaxas DN.
\newblock Multispectral deep neural networks for pedestrian detection  (2016).

\bibitem[{Gonz{\'a}lez et~al.(2016)Gonz{\'a}lez, Fang, Socarras, Serrat,
  V{\'a}zquez, Xu et~al.}]{gonzalez2016}
Gonz{\'a}lez A, Fang Z, Socarras Y, Serrat J, V{\'a}zquez D, Xu J, et~al.
\newblock Pedestrian detection at day/night time with visible and fir cameras:
  A comparison.
\newblock {\em Sensors\/} {\bf 16} (2016) 820.

\bibitem[{Zhang et~al.(2020)Zhang, Yang, NanLi, and Yu}]{yangmoyuan2020}
Zhang Y, Yang M, NanLi, Yu Z.
\newblock Analysis-synthesis dictionary pair learning and patch saliency
  measure for image fusion.
\newblock {\em Signal Processing\/} {\bf 167} (2020) 107327.

\bibitem[{Liu et~al.(2020)Liu, Wang, Cheng, Li, and Chen}]{liu2020}
Liu Y, Wang L, Cheng J, Li C, Chen X.
\newblock Multi-focus image fusion: A survey of the state of the art.
\newblock {\em Information Fusion\/} {\bf 64} (2020) 71--91.

\bibitem[{Li et~al.(2018{\natexlab{a}})Li, He, Tao, Tang, and
  Wang}]{li2018joint}
Li H, He X, Tao D, Tang Y, Wang R.
\newblock Joint medical image fusion, denoising and enhancement via
  discriminative low-rank sparse dictionaries learning.
\newblock {\em Pattern Recognition\/} {\bf 79} (2018{\natexlab{a}}) 130--146.

\bibitem[{Li et~al.(2020)Li, Wang, Yang, Wang, Li, and
  Tao}]{li2020discriminative}
Li H, Wang Y, Yang Z, Wang R, Li X, Tao D.
\newblock Discriminative dictionary learning-based multiple component
  decomposition for detail-preserving noisy image fusion.
\newblock {\em IEEE Transactions on Instrumentation and Measurement\/} {\bf 69}
  (2020) 1082--1102.

\bibitem[{Xie et~al.(2021)Xie, Wang, and Zhang}]{wangjiaxin2021}
Xie M, Wang J, Zhang Y.
\newblock A unified framework for damaged image fusion and completion based on
  low-rank and sparse decomposition.
\newblock {\em Signal Processing: Image Communication\/} {\bf 29} (2021)
  116400.

\bibitem[{Wang et~al.(2022{\natexlab{a}})Wang, Huang, Li, Qi, Tao, and
  Yu}]{huangbochun}
Wang S, Huang B, Li H, Qi G, Tao D, Yu Z.
\newblock Key point-aware occlusion suppression and semantic alignment for
  occluded person re-identification.
\newblock {\em Information Sciences\/}  (2022{\natexlab{a}}).

\bibitem[{Zhu et~al.(2021)Zhu, Luo, Chen, Qi, Mazur, Zhong et~al.}]{zhuzhipin}
Zhu Z, Luo Y, Chen S, Qi G, Mazur N, Zhong C, et~al.
\newblock Camera style transformation with preserved self-similarity and
  domain-dissimilarity in unsupervised person re-identification.
\newblock {\em Journal of Visual Communication and Image Representation\/} {\bf
  80} (2021) 103303.

\bibitem[{Yang et~al.(2022)Yang, Qian, Zhu, Wang, and Yang}]{yang2022}
Yang X, Qian Y, Zhu H, Wang C, Yang M.
\newblock Baanet: Learning bi-directional adaptive attention gates for
  multispectral pedestrian detection.
\newblock {\em 2022 International Conference on Robotics and Automation
  (ICRA)\/} (2022), 2920--2926.
\newblock \doi{10.1109/ICRA46639.2022.9811999}.

\bibitem[{Guan et~al.(2019)Guan, Cao, Yang, Cao, and Yang}]{guan2019}
Guan D, Cao Y, Yang J, Cao Y, Yang MY.
\newblock Fusion of multispectral data through illumination-aware deep neural
  networks for pedestrian detection.
\newblock {\em Information Fusion\/} {\bf 50} (2019) 148--157.

\bibitem[{Zhou et~al.(2020)Zhou, Chen, and Cao}]{zhou2020}
Zhou K, Chen L, Cao X.
\newblock Improving multispectral pedestrian detection by addressing modality
  imbalance problems.
\newblock {\em European Conference on Computer Vision\/} (Springer) (2020),
  787--803.

\bibitem[{Li et~al.(2022{\natexlab{c}})Li, Zhang, Hu, Fu, and Zhu}]{li2022}
Li Q, Zhang C, Hu Q, Fu H, Zhu P.
\newblock Confidence-aware fusion using dempster-shafer theory for
  multispectral pedestrian detection.
\newblock {\em IEEE Transactions on Multimedia\/}  (2022{\natexlab{c}}).

\bibitem[{Zhang et~al.(2020b)Zhang, Fromont, Lefevre, and Avignon}]{zhang2020b}
Zhang H, Fromont E, Lefevre S, Avignon B.
\newblock Multispectral fusion for object detection with cyclic fuse-and-refine
  blocks.
\newblock {\em 2020 IEEE International Conference on Image Processing (ICIP)\/}
  (2020b), 276--280.
\newblock \doi{10.1109/ICIP40778.2020.9191080}.

\bibitem[{Dollár et~al.(2014)Dollár, Appel, Belongie, and
  Perona}]{dollar2014}
Dollár P, Appel R, Belongie S, Perona P.
\newblock Fast feature pyramids for object detection.
\newblock {\em IEEE Transactions on Pattern Analysis and Machine
  Intelligence\/} {\bf 36} (2014) 1532--1545.
\newblock \doi{10.1109/TPAMI.2014.2300479}.

\bibitem[{Dalal and Triggs(2005)}]{dalal2015}
Dalal N, Triggs B.
\newblock Histograms of oriented gradients for human detection.
\newblock {\em 2005 IEEE Computer Society Conference on Computer Vision and
  Pattern Recognition (CVPR'05)\/} (2005), vol.~1, 886--893 vol. 1.
\newblock \doi{10.1109/CVPR.2005.177}.

\bibitem[{König et~al.(2017)König, Adam, Jarvers, Layher, Neumann, and
  Teutsch}]{konig2017}
König D, Adam M, Jarvers C, Layher G, Neumann H, Teutsch M.
\newblock Fully convolutional region proposal networks for multispectral person
  detection.
\newblock {\em 2017 IEEE Conference on Computer Vision and Pattern Recognition
  Workshops (CVPRW)\/} (2017), 243--250.
\newblock \doi{10.1109/CVPRW.2017.36}.

\bibitem[{Ren et~al.(2017)Ren, He, Girshick, and Sun}]{ren2017}
Ren S, He K, Girshick R, Sun J.
\newblock Faster r-cnn: Towards real-time object detection with region proposal
  networks.
\newblock {\em IEEE Transactions on Pattern Analysis and Machine
  Intelligence\/} {\bf 39} (2017) 1137--1149.
\newblock \doi{10.1109/TPAMI.2016.2577031}.

\bibitem[{Kim et~al.(2021a)Kim, Park, and Kim}]{kim2021a}
Kim J, Park I, Kim S.
\newblock A fusion framework for multi-spectral pedestrian detection using
  efficientdet.
\newblock {\em 2021 21st International Conference on Control, Automation and
  Systems (ICCAS)\/} (2021a), 1111--1113.
\newblock \doi{10.23919/ICCAS52745.2021.9650057}.

\bibitem[{Zhang et~al.(2019a)Zhang, Zhu, Chen, Yang, Lei, and Liu}]{zhang2019a}
Zhang L, Zhu X, Chen X, Yang X, Lei Z, Liu Z.
\newblock Weakly aligned cross-modal learning for multispectral pedestrian
  detection.
\newblock {\em 2019 IEEE/CVF International Conference on Computer Vision
  (ICCV)\/} (2019a), 5126--5136.
\newblock \doi{10.1109/ICCV.2019.00523}.

\bibitem[{Kim et~al.(2021b)Kim, Kim, Kim, Kim, and Choi}]{kim2021b}
Kim J, Kim H, Kim T, Kim N, Choi Y.
\newblock Mlpd: Multi-label pedestrian detector in multispectral domain.
\newblock {\em IEEE Robotics and Automation Letters\/} {\bf 6} (2021b)
  7846--7853.
\newblock \doi{10.1109/LRA.2021.3099870}.

\bibitem[{Kim et~al.(2022)Kim, Park, and Ro}]{kim2022}
Kim JU, Park S, Ro YM.
\newblock Uncertainty-guided cross-modal learning for robust multispectral
  pedestrian detection.
\newblock {\em IEEE Transactions on Circuits and Systems for Video
  Technology\/} {\bf 32} (2022) 1510--1523.
\newblock \doi{10.1109/TCSVT.2021.3076466}.

\bibitem[{Vaswani et~al.(2017)Vaswani, Shazeer, Parmar, Uszkoreit, Jones, Gomez
  et~al.}]{atention}
Vaswani A, Shazeer N, Parmar N, Uszkoreit J, Jones L, Gomez AN, et~al.
\newblock Attention is all you need.
\newblock {\em Proceedings of the 31st International Conference on Neural
  Information Processing Systems\/} (Red Hook, NY, USA: Curran Associates Inc.)
  (2017), NIPS'17, 6000–6010.

\bibitem[{Li et~al.(2020a)Li, Zou, Li, Zhao, and Gao}]{li2020a}
Li S, Zou C, Li Y, Zhao X, Gao Y.
\newblock Attention-based multi-modal fusion network for semantic scene
  completion.
\newblock {\em Proceedings of the AAAI Conference on Artificial Intelligence\/}
  (2020a), vol.~34, 11402--11409.

\bibitem[{Li et~al.(2020b)Li, Zhou, and Ren}]{li2020b}
Li B, Zhou Y, Ren H.
\newblock Image emotion caption based on visual attention mechanisms.
\newblock {\em 2020 IEEE 6th International Conference on Computer and
  Communications (ICCC)\/} (IEEE) (2020b), 1456--1460.

\bibitem[{Xiao et~al.(2022)Xiao, Zhang, Wang, Li, and Jin}]{xiaowanxin2022}
Xiao W, Zhang Y, Wang H, Li F, Jin H.
\newblock Heterogeneous knowledge distillation for simultaneous
  infrared-visible image fusion and super-resolution.
\newblock {\em IEEE Transactions on Instrumentation and Measurement\/} {\bf 71}
  (2022) 5004015.

\bibitem[{Li et~al.(2021c)Li, Cen, Liu, Chen, and Yu}]{cenyueliang2021}
Li H, Cen Y, Liu Y, Chen X, Yu Z.
\newblock Different input resolutions and arbitrary output resolution: A meta
  learning-based deep framework for infrared and visible image fusion.
\newblock {\em IEEE Transactions on Image Processing\/} {\bf 30} (2021c)
  4070--4083.

\bibitem[{Li et~al.(2022{\natexlab{d}})Li, Gao, Zhang, Xie, and
  Yu}]{gaojirui2022}
Li H, Gao J, Zhang Y, Xie M, Yu Z.
\newblock Haze transfer and feature aggregation network for real-world single
  image dehazing.
\newblock {\em Knowledge-Based Systems\/} {\bf 251} (2022{\natexlab{d}})
  109309.

\bibitem[{Xu et~al.(2021)Xu, Fu, Liu, and Li}]{xu2021}
Xu M, Fu P, Liu B, Li J.
\newblock Multi-stream attention-aware graph convolution network for video
  salient object detection.
\newblock {\em IEEE Transactions on Image Processing\/} {\bf 30} (2021)
  4183--4197.
\newblock \doi{10.1109/TIP.2021.3070200}.

\bibitem[{Li et~al.(2022{\natexlab{e}})Li, Xu, Li, and Yu}]{xukaixiong2022}
Li H, Xu K, Li J, Yu Z.
\newblock Dual-stream reciprocal disentanglement learning for domain adaptation
  person re-identification.
\newblock {\em Knowledge-Based Systems\/} {\bf 251} (2022{\natexlab{e}})
  109315.

\bibitem[{Zhang et~al.(2022)Zhang, Wang, Li, and Li}]{ACMMM2021}
Zhang Y, Wang Y, Li H, Li S.
\newblock Cross-compatible embedding and semantic consistent feature
  construction for sketch re-identification.
\newblock {\em Proceedings of the 30th ACM International Conference on
  Multimedia\/} (New York, NY, USA: Association for Computing Machinery)
  (2022), MM '22, 3347--3355.
\newblock \doi{10.1145/3503161.3548224}.

\bibitem[{Wang et~al.(2022{\natexlab{b}})Wang, Qi, Li, Chai, and
  Li}]{wangyiming}
Wang Y, Qi G, Li S, Chai Y, Li H.
\newblock Body part-level domain alignment for domain-adaptive person
  re-identification with transformer framework.
\newblock {\em IEEE Transactions on Information Forensics and Security\/} {\bf
  17} (2022{\natexlab{b}}) 3321--3334.
\newblock \doi{10.1109/TIFS.2022.3207893}.

\bibitem[{Hu et~al.(2020)Hu, Shen, Albanie, Sun, and Wu}]{hu2020}
Hu J, Shen L, Albanie S, Sun G, Wu E.
\newblock Squeeze-and-excitation networks.
\newblock {\em IEEE Transactions on Pattern Analysis and Machine
  Intelligence\/} {\bf 42} (2020) 2011--2023.
\newblock \doi{10.1109/TPAMI.2019.2913372}.

\bibitem[{Li et~al.(2019a)Li, Wang, Hu, and Yang}]{li2019a}
Li X, Wang W, Hu X, Yang J.
\newblock Selective kernel networks.
\newblock {\em 2019 IEEE/CVF Conference on Computer Vision and Pattern
  Recognition (CVPR)\/} (2019a), 510--519.
\newblock \doi{10.1109/CVPR.2019.00060}.

\bibitem[{Dai et~al.(2021)Dai, Gieseke, Oehmcke, Wu, and Barnard}]{dai2021}
Dai Y, Gieseke F, Oehmcke S, Wu Y, Barnard K.
\newblock Attentional feature fusion.
\newblock {\em 2021 IEEE Winter Conference on Applications of Computer Vision
  (WACV)\/} (2021), 3559--3568.
\newblock \doi{10.1109/WACV48630.2021.00360}.

\bibitem[{He et~al.(2016)He, Zhang, Ren, and Sun}]{he2016}
He K, Zhang X, Ren S, Sun J.
\newblock Deep residual learning for image recognition.
\newblock {\em 2016 IEEE Conference on Computer Vision and Pattern Recognition
  (CVPR)\/} (2016), 770--778.
\newblock \doi{10.1109/CVPR.2016.90}.

\bibitem[{Woo et~al.(2018)Woo, Park, Lee, and Kweon}]{woo2018}
Woo S, Park J, Lee JY, Kweon IS.
\newblock Cbam: Convolutional block attention module.
\newblock {\em Proceedings of the European conference on computer vision
  (ECCV)\/} (2018), 3--19.

\bibitem[{Girshick(2015)}]{girshick2015}
Girshick R.
\newblock Fast r-cnn.
\newblock {\em 2015 IEEE International Conference on Computer Vision (ICCV)\/}
  (2015), 1440--1448.
\newblock \doi{10.1109/ICCV.2015.169}.

\bibitem[{Li et~al.(2018{\natexlab{b}})Li, Song, Tong, and Tang}]{li2018}
Li C, Song D, Tong R, Tang M.
\newblock Multispectral pedestrian detection via simultaneous detection and
  segmentation.
\newblock {\em arXiv preprint arXiv:1808.04818\/}  (2018{\natexlab{b}}).

\bibitem[{Dollar et~al.(2012)Dollar, Wojek, Schiele, and Perona}]{dollar2012}
Dollar P, Wojek C, Schiele B, Perona P.
\newblock Pedestrian detection: An evaluation of the state of the art.
\newblock {\em IEEE Transactions on Pattern Analysis and Machine
  Intelligence\/} {\bf 34} (2012) 743--761.
\newblock \doi{10.1109/TPAMI.2011.155}.

\bibitem[{Xu et~al.(2017)Xu, Ouyang, Ricci, Wang, and Sebe}]{xu2017}
Xu D, Ouyang W, Ricci E, Wang X, Sebe N.
\newblock Learning cross-modal deep representations for robust pedestrian
  detection.
\newblock {\em 2017 IEEE Conference on Computer Vision and Pattern Recognition
  (CVPR)\/} (2017), 4236--4244.
\newblock \doi{10.1109/CVPR.2017.451}.

\bibitem[{Zhang et~al.(2019b)Zhang, Liu, Zhang, Yang, Qiao, Huang
  et~al.}]{zhang2019b}
Zhang L, Liu Z, Zhang S, Yang X, Qiao H, Huang K, et~al.
\newblock Cross-modality interactive attention network for multispectral
  pedestrian detection.
\newblock {\em Information Fusion\/} {\bf 50} (2019b) 20--29.

\bibitem[{Li et~al.(2019b)Li, Song, Tong, and Tang}]{li2019b}
Li C, Song D, Tong R, Tang M.
\newblock Illumination-aware faster r-cnn for robust multispectral pedestrian
  detection.
\newblock {\em Pattern Recognition\/} {\bf 85} (2019b) 161--171.

\bibitem[{Zhang et~al.(2020a)Zhang, Yin, Nie, and Huang}]{zhang2020a}
Zhang Y, Yin Z, Nie L, Huang S.
\newblock Attention based multi-layer fusion of multispectral images for
  pedestrian detection.
\newblock {\em IEEE Access\/} {\bf 8} (2020a) 165071--165084.
\newblock \doi{10.1109/ACCESS.2020.3022623}.

\bibitem[{Zhuang et~al.(2022)Zhuang, Pu, Hu, and Wang}]{zhuang2022}
Zhuang Y, Pu Z, Hu J, Wang Y.
\newblock Illumination and temperature-aware multispectral networks for
  edge-computing-enabled pedestrian detection.
\newblock {\em IEEE Transactions on Network Science and Engineering\/} {\bf 9}
  (2022) 1282--1295.
\newblock \doi{10.1109/TNSE.2021.3139335}.

\bibitem[{Liu et~al.(2022)Liu, Lam, Zhao, and Qiu}]{liu2022}
Liu T, Lam KM, Zhao R, Qiu G.
\newblock Deep cross-modal representation learning and distillation for
  illumination-invariant pedestrian detection.
\newblock {\em IEEE Transactions on Circuits and Systems for Video
  Technology\/} {\bf 32} (2022) 315--329.
\newblock \doi{10.1109/TCSVT.2021.3060162}.

\end{thebibliography}
	
	
	\section*{Figure captions}

\end{document}